# De-identifying Australian Hospital Discharge Summaries: An End-to-End Framework using Ensemble of Deep Learning Models

**Leibo Liu[a], Oscar Perez-Concha[a], Anthony Nguyen[b], Vicki Bennett[c], Louisa Jorm[a]**

[a]Centre for Big Data Research in Health, University of New South Wales, Sydney New South Wales 2052, Australia

[b]The Australian e-Health Research Centre, CSIRO, Brisbane, Queensland 4029, Australia

[c]Metadata and METeOR Unit, Australian Institute of Health and Welfare, Canberra Australian Capital Territory 2601, Australia

[*]**Corresponding Author**: Leibo Liu, Centre for Big Data Research in Health, Level 2, AGSM Building (G27) University of New South Wales Sydney, Kensington, New South Wales, 2052 Australia (leibo.liu@student.unsw.edu.au, phone +61 2 9065 7847)

**Abstract**

Electronic Medical Records (EMRs) contain clinical narrative text that is of great potential value to medical researchers. However, this information is mixed with Personally Identifiable Information (PII) that presents risks to patient and clinician confidentiality. This paper presents an end-to-end de-identification framework to automatically remove PII from Australian hospital discharge summaries. Our corpus included 600 hospital discharge summaries which were extracted from the EMRs of two principal referral hospitals in Sydney, Australia. Our end-to-end de-identification framework consists of three components: 1) Annotation: labelling of PII in the 600 hospital discharge summaries using five pre-defined categories: person, address, date of birth, individual identification number, phone/fax number; 2) Modelling: training six named entity recognition (NER) deep learning base-models on balanced and imbalanced datasets; and evaluating ensembles that combine all six base-models, the three base-models with the best F1 scores and the three base-models with the best recall scores respectively, using token-level majority voting and stacking methods; and 3) De-identification: removing PII from the hospital discharge summaries. Our results showed that the ensemble model combined using the stacking Support Vector Machine (SVM) method on the three base-models with the best F1 scores achieved excellent results with a F1 score of 99.16% on the test set of our corpus. We also evaluated the robustness of our modelling component on the 2014 i2b2 de-identification dataset. Our ensemble model, which uses the token-level majority voting method on all six base-models, achieved the highest F1 score of 96.24% at strict entity matching and the highest F1 score of 98.64% at binary token-level matching compared to two state-of-the-art methods. The end-to-end framework provides a robust solution to de-identifying clinical narrative corpuses safely. It can easily be applied to any kind of clinical narrative documents.

# 1. Introduction

There is currently intense interest in using narrative free text from Electronic Medical Records (EMRs) to develop predictive algorithms for patient stratification and personalised health care. However, sharing of and access to this clinical text for research purposes are severely constrained because it includes information that may identify specific individuals, known as Personally Identifiable Information (PII) [1]. Development of robust and scalable yet resource-efficient methods for de-identifying clinical narrative text is a priority for making these valuable data more available to researchers [2] to generate evidence that will deliver benefits to both health care providers and patients [3].

In Australia, there is no clear definition of the term "de-identified health data" or agreed standards for de-identification, even though a strong health privacy regulatory environment is in place [4]. In contrast, the Health Insurance Portability and Accountability Act (HIPAA) which was passed in United States in 1996 provides two de-identification methods: 1) The "Expert Determination" method: qualified experts analyse and determine the risk of re-identifying an individual who is a subject of the information; and 2) The "Safe Harbor" method: this method requires the removal of 18 types of identifiers (see Table S1), including names, geographic subdivisions, telephone numbers, medical record numbers and so on [5].

Manual de-identification of a large clinical corpus is a time-consuming and challenging task [6, 7]. Some studies have explored the possibility of using Natural Language Processing (NLP) and deep learning techniques for automatic de-identification relevant to the "Safe Harbor" method [8-10]. Three de-identification challenges, organized by the Informatics for Integrating Biology and the Bedside (i2b2) organization in 2006 [11] and 2014 [12], and Centres of Excellence in Genomic Science (CEGS) and Neuropsychiatric Genome-Scale and RDOC Individualized Domains (N-GRID) in 2016 [13], facilitated innovation in de-identification methods based on NLP and proved the efficiency of automatic de-identification. However, these challenges involved building models and comparing results using a specific annotated dataset. There is a need for end-to-end de-identification solutions which can be easily applied to raw clinical narrative documents, especially within the internet access-restricted environments of hospitals.

In this paper, we propose an end-to-end de-identification framework with application to Australian hospital discharge summaries. This paper describes the three components of the end-to-end framework: 1) Annotation: a web-based annotation tool to help our human annotators tag the PII entities in the raw hospital discharge summaries; 2) Modelling: training and evaluating ensembles of different named entity recognition (NER) deep learning models to identify PII in hospital discharge summaries; 3) De-identification: automatic tag and removal of PII entities from hospital discharge summaries.

## 2. Related work

### 2.1 Model architectures

In the past decade, several methods for automatic de-identification have been developed and evaluated [9]. Rule-based methods such as that proposed by Menger et al [3] do not need pre-annotated corpora and the models can be improved easily by adding or modifying rules. However, the rules used for de-identification need to be manually built by experienced domain experts and are very difficult to generalize to other corpora. Other work has focused on machine learning-based models [6, 9, 10] and hybrid systems that combine rule-based and machine learning techniques [7].

More recently, ensemble-based methods have been shown to outperform other approaches. Kim et al. [14] combined twelve individual models which were trained or developed using rule-based and machine learning methods with three different ensemble methods: voting, decision template method [15] and stacked generalization [16]. The individual and ensemble models were evaluated on the 2014 i2b2 test set [17]. The stacked ensemble model achieved the highest F1 score of 95.73%. Murugadoss et al. [18] presented a de-identification tool through ensemble learning of multiple binary entity fine-tuned Bidirectional Encoder Representations from Transformers (BERT) [19] models and rule-based methods. Each BERT model was only used to predict one of four PII entity classes (Name, Location, Age, Organization). The ensemble learning method obtained a F1 score of 98.5% on the 2014 i2b2 test set. As ensemble models achieved the best performances, we used these two ensemble methods as baselines in our study.

### 2.2 Customisable versus off-the-shelf systems

How well de-identification solutions generalize to previously unseen data is a critical issue given the variability in the narrative clinical text captured across settings [13, 20]. A challenge organized by Centres of Excellence in Genomic Science Neuropsychiatric Genome-Scale and Research Domain Criteria Individualized Domains (CEGS N-GRID) in 2016 contained a de-identification task which consisted of two subtasks [13]: 1) To use existing de-identification systems, which were not trained or modified, on the Research Domain Criteria (RDoC) test data to assess how well the systems generalized to the newly unseen data; 2) To develop new models using the RDoC. Teams that participated in the first subtask achieved median and maximum F1 scores of 62.9% and 79.9%, compared to 82.2% and 91.4%, respectively, for teams completing the second task, demonstrating that the performance of the de-identification systems could be improved with customized training. Similarly, a significant improvement in F1 score (from 85.68% to 92.88%) was achieved when de-identifying clinical notes from University of Florida Health using a customized model, compared to using a deep learning model trained using the 2014 i2b2 corpus [10]. Hartman et al. [20] investigated and assessed performance of four different levels of customization scenarios from off-the-shelf to fully customized on six datasets.

The fully customized models which were trained on specific datasets consistently outperformed the off-the-shelf models which were pretrained on other datasets.

Heider et al. [21] compared and evaluated three off-the-shelf de-identification systems: 1) Amazon Comprehend Medical [22], which uses machine learning techniques to extract relevant medical information from unstructured text and can identify the PII described in the HIPAA's Safe Harbor method; 2) Clinacuity CliniDeID [23], an automatic clinical text de-identification tool which can accurately de-identify and replace PII with realistic surrogates in clinical notes according to the HIPAA's Safe Harbor method; 3) NLM Scrubber [24], a free, HIPAA-compliant clinical text de-identification tool which can be downloaded and run on your local computer. They reported that CliniDeID had the best F1 scores compared to the other two tools, while NLM Scrubber achieved the fastest speed of about 13.69 secs per 10k characters. However, these off-the-shelf tools cannot be trained and tailored to specific datasets within an internet access-restricted environment such as hospital networks. As a result, their performance on these datasets cannot be improved.

## 2.3 End-to-end solutions

In order to efficiently deploy customisable de-identification solutions, tools are required to facilitate training dataset annotation, model training, and redaction of identifier text. An integrated framework, the Health Information DE-identification (HIDE) was proposed by Gardner et al. [25] for de-identifying health information including both structured and unstructured data. HIDE relied on Naïve Bayes conditional random field (CRF) algorithms to train the NER models and provided an environment for tagging, classifying, and retagging of PII. The MITRE Identification Scrubber Toolkit (MIST) implemented by Aberdeen et al. [26] provided a comprehensive environment to rapidly develop a de-identification tool that included a web-based graphical annotation tool, a training module, a tagging module, and a redaction module. MIST trained the NER models based on CRF algorithm and achieved an overall F1 of 96.5% on the 2006 i2b2 dataset.

## 3. Materials and methods

In this section, we describe the corpus that we used and the three components of our proposed end-to-end de-identification framework.

## 3.1 Corpus and PII categories

We used hospital discharge summaries extracted from the EMRs of two principal referral hospitals in Sydney, Australia according to the following inclusion criteria:

1) Patients were >=18 years old;

2) Hospital stay was longer than 24 hours; and

3) Admission was for non-maternity related reasons.

A total of 3,554 discharge summaries were included in our corpus. We first built our annotated datasets that were used to train our NER models. Considering the time and effort of manual annotation of PII, 600 discharge summaries were randomly chosen. They were split into three sets: 400 discharge summaries as training set, 100 discharge summaries as development set and 100 discharge summaries as test set. We trained our NER base-models using the training set and tuned the model parameters on the development set. The test set was used to evaluate the unbiased performance of the different models.

In Australia, there is no agreed definition of a "de-identified" dataset, rather a comprehensive risk management approach that considers treatment of the data alongside assessment of the benefits of the project, credentials of the analysts, security of the analysis environment and methods for assessing disclosure risks of outputs is used [27]. Based on the PII categories of the HIPAA Safe Harbor de-identification method, Australian guidelines for de-identification [28], and the content of discharge summaries, we grouped the PII in the discharge summaries to five categories (see Table 1). We (1) merged telephone numbers and fax numbers as PHONE category; (2) merged all ID numbers listed in HIPAA's categories (G-M, R) such as social security numbers, medical record numbers (MRN), health plan beneficiary numbers and the like as IDN category; (3) disregarded the email, Web Universal Resource Locators (URLs), Internet Protocol (IP) addresses, images and finger/voice prints because there is no such information collected in our corpus; (4) only focused on date of birth as DOB category and kept all the other dates. The ethically approved uses of our de-identified datasets required date information except date of birth to be kept.

Table 1. Five PII Categories in our corpus.

| | PII Category | Description |
|---|---|---|
| 1 | PERSON | Patient names, Doctor names, Family member names and other names |
| 2 | ADDRESS | Patient address, General Practitioner address |
| 3 | DOB | Date of Birth |
| 4 | IDN | Medical Record Number (MRN), Financial Identification Number (FIN), Doctor pager number |
| 5 | PHONE | Phone number, fax number |

### 3.2 Framework components

The end-to-end de-identification framework includes three components: Annotation, Modelling and De-identification. Fig. 1 describes the framework architecture and workflow of de-identification. Human annotators utilize the Annotation component to tag pre-defined PII entities in the discharge summaries. The annotated discharge summaries are input into the Modelling component to train and select the best ensemble model, which is used for de-identification in the De-Identification component.

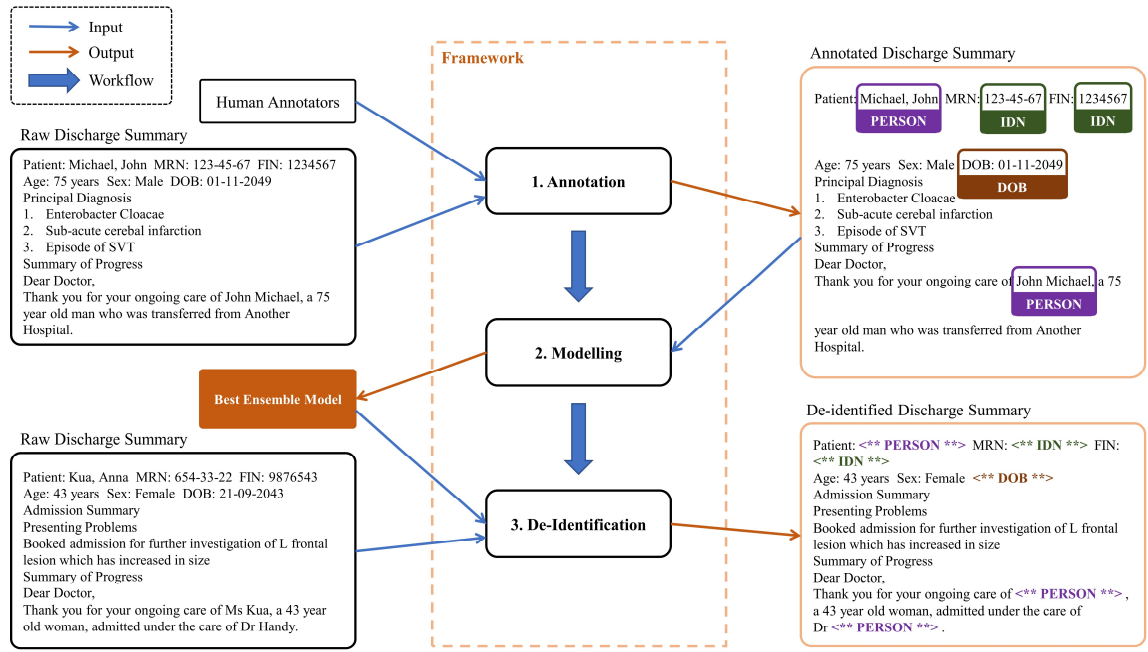

**Fig. 1.** The architecture and workflow of the end-to-end de-identification framework.

### 3.2.1 Annotation

This component of the framework receives the 600 selected discharge summaries as the input and outputs the annotated discharge summaries, through an annotation tool that assists human annotators to tag the discharge summaries.

There are some existing natural language annotation tools such as Multi-document Annotation Environment (MAE) [29] and Brat rapid annotation tool (BRAT) [30]. MAE is a multi-purpose annotation tool which had been used for semantic annotations and discourse relation detection. It is written as a Java standalone application. As a result, it can only be deployed on a local computer and accessed by the local users. It is very difficult to maintain and not suitable for team working. BRAT is a web-based annotation tool with high-quality visualisation and intuitive annotation interface. However, the latest version released in 2012 was developed in Python version 2.5 which is not compatible with Python version 3.0+ which we used in our framework. The annotation results of BRAT can only be exported as a list of entities with index ranges. A data format transformation needs to be done before training the NER models.

To tackle the problems of the existing annotation tools, we developed a web-based annotation tool (WAT) to assist the annotation process. The WAT was developed using React (a JavaScript library) [31] and Flask [32] based on Python 3.7+. It loads contents of discharge summaries and presents a user-friendly interface for rapid annotation of PII. Annotated PII entities are colour-coded with an intuitive workflow for adding, editing, and deleting text annotations (see Fig. S1). After the human annotator confirms the completion of annotation of one discharge summary, a save action can be processed to store the annotated content of the discharge summary using Beginning-Inside-Outside (BIO) format [33]. The BIO format data will form the input for the downstream modelling component.

To build the annotated datasets, the five categories of PII were tagged manually by human annotators in the 600 discharge summaries using WAT. Two annotators annotated the test set (100 discharge summaries). To validate the quality of the annotation, the inter-annotator agreement (IAA) was measured on the test set. Cohen's Kappa has been proven not to be a good IAA measure for NER tasks because the number of negative cases for NER are undefined [34]. To resolve this issue, token-level Cohen's Kappa was proposed as a solution [35, 36]. However, it does not reflect the IAA very well due to entity versus token span differences, and the extremely unbalanced data (NER tag "O") which will overestimate the Kappa score. A solution which calculates the Kappa score only using annotated tokens was proposed by Brandsen et al. [37], although this method will underestimate the IAA. Due to these issues with Cohen's Kappa, the F1 score has been widely used to measure IAA in NER tasks [37, 38]. We reported on both the F1 score and two types of token-level Cohen's Kappa in Table 2. When calculating F1 score for IAA, the annotation of one reviewer was used as the ground truth and the annotation of the other reviewer as the predictions of a system. The F1 score remained the same when switching the roles of the two annotators. Most of the disagreements were because incorrect boundaries e.g., one annotator tagged "1234-56-7" as IDN entity while the other annotator only tagged "1234-56-" which missed the last number. Given that the Cohen's Kappa scores and the F1 score were all high, we were confident to annotate the rest of discharge summaries using only one annotator. This was part of our experimental design, allowing us to investigate whether this annotation methodology would build performant NER models in a resource-limited domain. A third annotator reviewed and corrected any discrepancies between the two annotated results from the test set. The test set as a gold standard set was used to estimate the unbiased performance of our end-to-end de-identification framework.

Table 2. Inter-annotator agreement (IAA) on the test set.

| IAA metric | Score |
|---|---|
| Cohen's Kappa on all tokens | 96.65 |
| Cohen's Kappa on annotated tokens only | 92.55 |
| F1 score (strict entity matching) | 94.74 (Overall)<br>F1 scores by category as below: |

| | | |
|---|---|---|
| | ADDRESS | 96.22 |
| | DOB | 95.58 |
| | PERSON | 94.88 |
| | PHONE | 97.17 |
| | IDN | 90.66 |

### 3.2.2 Modelling

The modelling component is the core and the most important part of our framework. It is fed the output of the upstream annotation component and outputs the best ensemble model to the downstream de-identification component.

There are multiple text lines in a discharge summary. Only 8,064 (11.33%) text lines in the training set and 2,250 (11.74%) in the development set had PII (see Table S2). Imbalanced data generally bias classifications to the majority classes (NER tag "O" in our corpus) and reduce the recall on minor classes. To investigate and address the effects of different proportions of text lines without PII, we created two different types (as shown in Fig. 2) of training and development sets, whereas the test set included all the text lines in the discharge summaries and was the same for both types:

1) Balanced dataset included all the text lines with PII and equal numbers of text lines without PII, randomly selected from the original training and development sets.

2) Imbalanced dataset included all the text lines in the original training and development sets, even if there were no PII in the text lines.

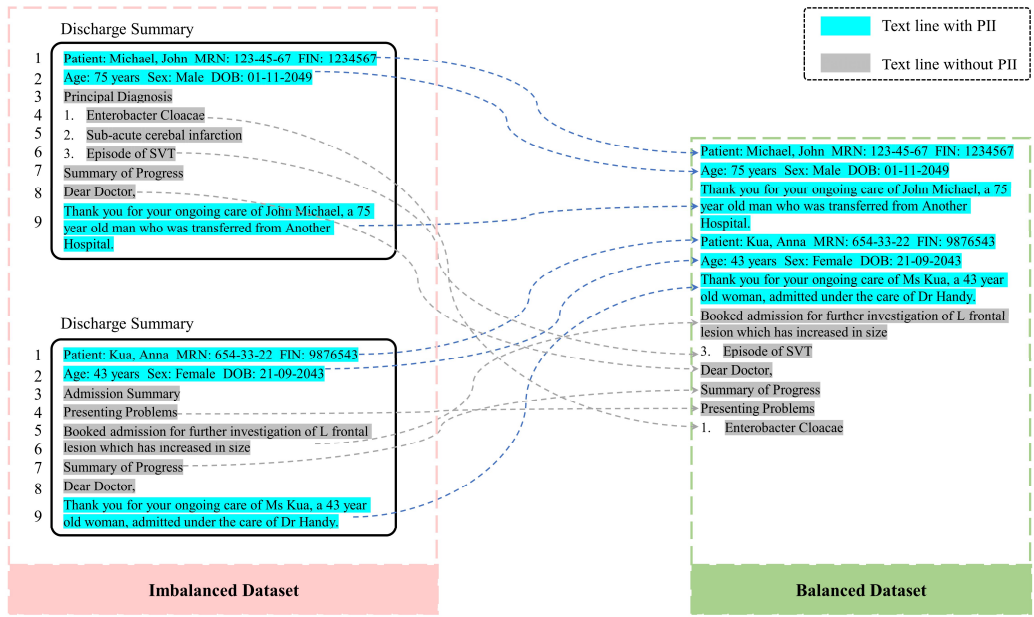

**Fig. 2.** An example of creating balanced and imbalanced datasets. Six text lines with PII are extracted and six text lines without PII are randomly selected from two discharge summaries to generate the balanced dataset.

From an NLP point of view, the de-identification task is a NER problem. The Bidirectional Long Short-Term Memory + Conditional Random Field (BiLSTM+CRF) architecture has been shown to achieve state-of-the-art (SOTA) or competitive performances in the clinical text de-identification task [10, 39, 40]. Two NLP libraries, Stanza [41] and Flair [42], make it easy and efficient to build BiLSTM+CRF NER models. Another popular deep learning architecture, the convolution neural network (CNN), has also been evaluated for NER tasks [43, 44, 45]. The spaCy [46] library provides pre-trained CNN models and can be easily fine-tuned for the de-identification task. In the modelling component, we fine-tuned two convolution neural network (CNN) NER base-models using spaCy and trained four BiLSTM+CRF NER base-models using Stanza and Flair on balanced and imbalanced datasets separately and ensembled the NER deep learning base-models using different methods (see Fig. 3). Word embeddings project words onto a continuous space to represent semantic and syntactic similarities between them [47]. The words need to be converted to a set of vectors before they are fed into NLP models. It is a crucial and first step when training an NLP model. There are many available word embeddings: Word2vec [48], Global Vectors for word representation (GloVe) [49], Embeddings from Language Models (ELMo) [50], contextualized embeddings from BERT and others. To combine the advantages of different word embeddings, we used different word embeddings for different NER base-models. spaCy provides some pre-trained pipelines which include multiple NLP components such as parser, tok2vec, ner and so on. We chose the pretrained spaCy pipeline 'en_core_web_lg' and appended our above-mentioned five PII categories into the label scheme of its NER CNN model. We performed transfer learning on our datasets using the pretrained spaCy pipeline to update the NER model parameters. We adopted the pre-trained Word2vec embedding vectors released from Computational Natural Language Learning (CoNLL) 2017 Shared Task to train the BiLSTM+CRF NER base-models using the Stanza library according to [40]. The concatenated word representations from all the 12 layers of the pre-trained BERT model 'bert-base-uncased' was extracted as the input features for training the BiLSTM+CRF NER base-models using Flair.

Hyperparameter optimization for the BiLSTM+CRF base-models was done using a search space of batch size (16, 32, 64), learning rate (0.1, 0.01, 0.001) and dropout (0, 0.1, 0.5). We utilized Bayesian optimization algorithm with the help of scikit-optimize library [51] to fine-tune the three hyperparameters. The learning rates in BiLSTM+CRF models were reduced gradually from the initial values to get the best models during the training processes. As mentioned above, fine-tuning the spaCy CNN models was achieved with transfer learning using its default hyperparameters.

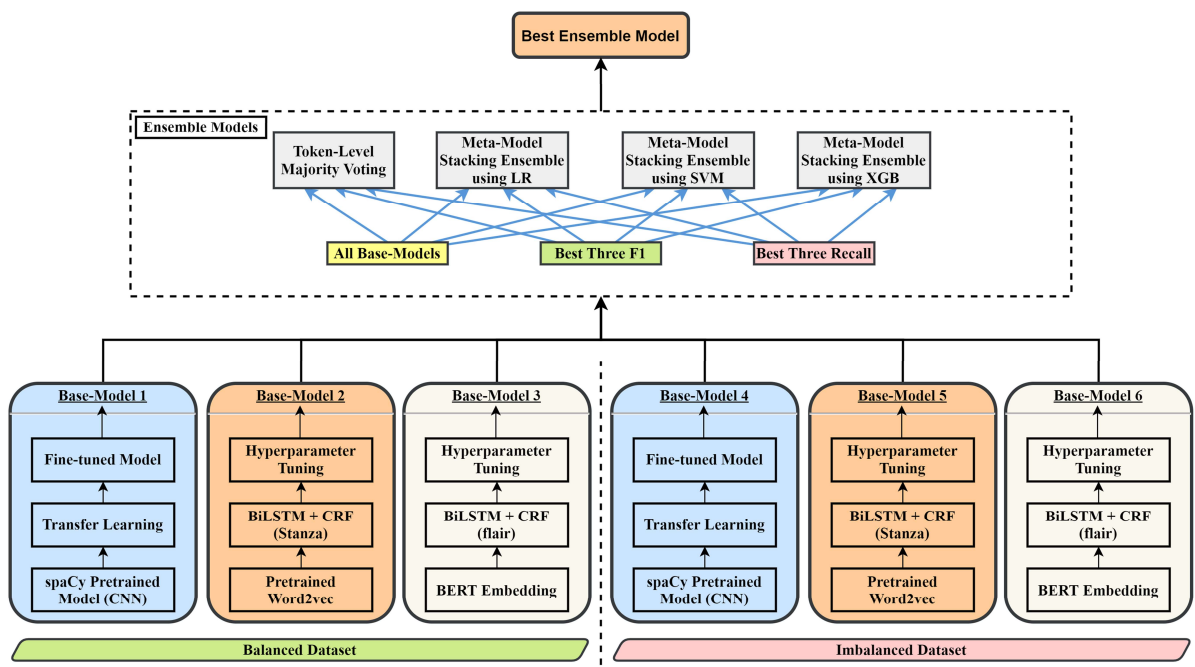

**Fig. 3**. Training process of modelling component. Six base-models were fine-tuned or trained on the balanced and imbalanced datasets. Multiple ensemble models were combined from all six base-models, the three models with the best F1 scores and the three models with the best recall scores respectively, using token-level majority voting and three different stacking algorithms. The ensemble model with the best F1 score were chosen as our final model for de-identification.

In order to achieve the best performance on the NER task, we used a stacking ensemble learning approach as ensemble-based methods have been shown to improve classification performances and outperform individual classification models [14]. We generated the ensemble models from three different groups of base-models that were trained on our datasets: 1) all six base-models; 2) the three base-models with the best F1 scores and 3) the three base-models with the best recall scores. We experimented with two types of ensemble methods: 1) majority voting and 2) stacking upon the three groups of base-models (see Fig. 3). As shown in Fig. S2, our voting ensemble compared the predictions of the base-models on token-level and voted the majority tag for each token. The tag from the base-model with the best F1 score was selected if no majority existed. We created three different stacking meta-models using logistic regression (LR), support vector machine (SVM) and extreme gradient boosting (XGB) algorithms. The predictions of the base-models on our development dataset were used as input to the meta-models which were trained on the development dataset. The ensemble model that outperformed other ensemble models on the test set was picked as our final NER model for the downstream de-identification component.

Model performance was evaluated using Precision, Recall and F1 score. The calculation of F1 score may differ according to the evaluation level (entity level - micro or category level - macro) and entity matching strictness (strict or relaxed) [9]. Strict entity matching, which was the primary evaluation metric for the 2014 i2b2 challenge [17], must exactly match the first and last offset of the gold standard entities. In this paper, all F1 scores are strict entity matching micro-F1 score unless specified otherwise.

Finally, in order to measure the robustness of our modelling component, we compared the performance of our models with two SOTA methods [14, 18] on the 2014 i2b2 dataset which consists of a training set of 521 clinical notes, a development set of 269 clinical notes and a test set of 514 clinical notes for the de-identification challenge [17]. Following our modelling process, both balanced and imbalanced datasets were created first according to the numbers of different types of text lines as shown in Table S3. Six base-models were fine-tuned or trained on both balanced and imbalance datasets and 12 ensemble models were combined upon three different groups of base-models using majority voting and stacking methods. Finally, the token-level majority voting ensemble model generated on all the base-models performed the best (see Table S4). For comparison with other methods, we also reported the F1 score of binary PII token-based matching which evaluated the performance based on the detection of PII tokens vs non-PII tokens [52].

### 3.2.3 De-Identification

The final component of the framework is to replace PII entities with surrogates. This component removed the PII entities which were predicted by the ensemble model generated by the modelling component and replaced these with pre-defined surrogates of "<** [ENTITYTYPENAME] **>" in the discharge summaries. For example, the sentence of "…Thank you for the care of Firstname Lastname, a 30-year-old man from..." will be de-identified to "…Thank you for the care of <**PERSON**>, a 30-year-old man from...".

### 4. Results

Our training dataset contains 10,536 PII entities and our development dataset contains 2,921 PII entities. There are 2,689 PII entities in the test set. Table 3 shows a summary of the PII numbers for each PII category in the training, development, and test sets. More than half of the PII entities (54%) are person names. IDN and PHONE categories appear in the corpus in nearly the same proportions (around 15%), with slightly fewer ADDRESS categories (11%). The DOB category has the least numbers, comprising only 5% of PII. The proportions of every category are quite similar among the different datasets, showing that the discharge summaries have similar content structures.

Table 3. Summary of the PII numbers in the datasets.

| Dataset | Discharge Summary | Text Lines with PII | PERSON | IDN | PHONE | ADDRESS | DOB | Total |
|---|---|---|---|---|---|---|---|---|
| Training | 400 | 8,064 | 5,724 | 1,624 | 1,511 | 1,201 | 476 | 10,536 |
| Development | 100 | 2,250 | 1,575 | 440 | 452 | 328 | 126 | 2,921 |
| Test | 100 | 2,060 | 1,451 | 426 | 380 | 308 | 124 | 2,689 |
| Total | 600 | 12,374 | 8,750 | 2,490 | 2,343 | 1,837 | 726 | 16,146 |

We completed the hyperparameter tuning on a CPU-based environment with 32GB memory and 8vCPU in a secure cloud computing infrastructure. The best hyperparameters for the BiLSTM+CRF base-models are shown in Table 4.

Table 4. Fine-tuned hyperparameters of the BiLSTM+CRF base-models.

| Dataset | Model | Batch Size | Learning Rate | Dropout |
|---|---|---|---|---|
| Balanced | BiLSTM+CRF (flair) | 16 | 0.1 | 0.5 |
| | BiLSTM+CRF (Stanza) | 16 | 0.1 | 0.5 |
| Imbalanced | BiLSTM+CRF (flair) | 64 | 0.1 | 0.5 |
| | BiLSTM+CRF (Stanza) | 16 | 0.1 | 0.5 |

Table 5 shows the evaluation results measured with individual base-models and all the ensemble meta-models on our test set. For the same base-model architecture, the base-model trained on the balanced dataset outperformed the one trained on the imbalanced dataset on the recall metric, while the imbalanced dataset improved the precision metric. The ensemble model combined using stacking SVM method on the three base-models with the best F1 scores produced the best F1 score of 99.16% among all the models. It obtained improved recall and F1 and a slightly reduced precision value, compared to the best base-model that had a F1 score of 99.05%.

Table 5. Evaluation results of individual base-models and ensemble models. The best results for each metric appear in bold. * indicates that the difference between the relevant model and the best model on the specific metric is significant (p<.05). The p-values are reported in Table S5.

| Model \ Dataset | BiLSTM+CRF (flair) | | | BiLSTM+CRF (Stanza) | | | CNN (spaCy) | | |
|---|---|---|---|---|---|---|---|---|---|
| | Precision | Recall | F1 | Precision | Recall | F1 | Precision | Recall | F1 |
| Balanced | 98.92* | 98.85* | 98.88* | 97.99 | 98.03* | 98.01* | 97.78 | 98.44* | 98.11* |
| Imbalanced | **99.62** | 98.48* | 99.05 | 98.90* | 97.32* | 98.11* | 98.83* | 97.66* | 98.24* |

| Ensemble Model | All Base-Models | | | Best Three F1 | | | Best Three Recall | | |
|---|---|---|---|---|---|---|---|---|---|
| | Precision | Recall | F1 | Precision | Recall | F1 | Precision | Recall | F1 |
| Majority Voting | 99.44 | 98.77 | 99.10 | 99.48* | 98.81 | 99.14 | 99.33* | **98.92** | 99.12 |
| Stacking LR | 98.99 | 98.10* | 98.54* | 99.36 | 98.62 | 98.99* | 99.47 | 98.21 | 98.84* |
| Stacking SVM | 99.10 | 98.44 | 98.77* | 99.55* | 98.77 | **99.16** | 99.48* | 98.74 | 99.10 |
| Stacking XGB | 99.28* | 98.07 | 98.67* | 98.99 | 98.25 | 98.62* | 99.33 | 98.59 | 98.95 |

We performed an error analysis for the best ensemble model on the test set. Most of the false positives and false negatives were in the PERSON category. Fig. 4 summarizes the count of false positives and false negatives on the test set.

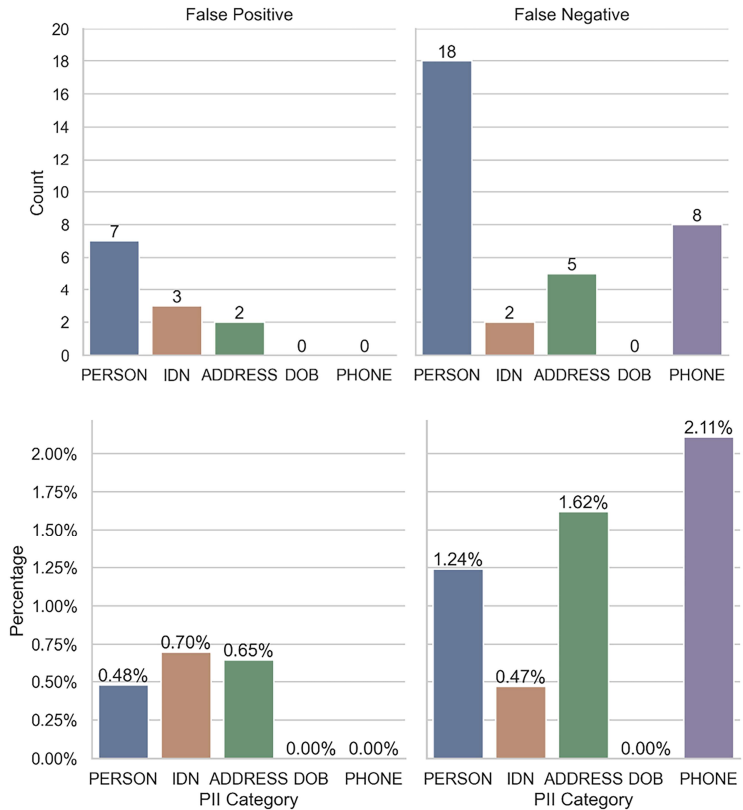

**Fig. 4.** Count and percentage of false positive and false negative by PII categories on the test set. There were 1451 PERSON entities, 426 IDN entities, 308 ADDRESS entities, 124 DOB entities and 380 PHONE entities on the test set, respectively.

A total of 7 entities were mistakenly recognized as PERSON by the model, falling into two categories:

1) Medical terminology (n=5). For example, 'IVOR LEWIS Esophagectomy' and 'Epley's manoeuvre'. 'IVOR LEWIS' and 'Epley's' were recognized incorrectly as PERSON entities by the model. In the sentence 'MRSA UTI (4/5/52) …', 'UTI' was tagged as PERSON mistakenly. UTI stands for urinary tract infection.

2) Boundary mismatch (n=2). The start or end indexes of the entities were not correct. For example, in the sentence 'Known to Dr Firstname Lastname's Room', the model tagged 'Firstname Lastname's Room' as a PERSON entity which mistakenly included 'Room' as part of the PERSON entity.

There were three false positive entities related to the IDN category. One of the three entities was a phone extension number which belonged to PHONE category and needed to be removed and the other two were only matching part of the entities. Two false positive entities in the ADDRESS category were the result of boundary mismatching.

A total of 18 PERSON entities were not identified by the model. These comprised five entities that were patient names, and 13 entities that were clinician names. The PERSON false negatives fell into three categories:

1) Boundary mismatch (n=2). Taking the sentence 'Dr Firstname Lastname Surgery' as an example, the model tagged one PERSON entity of 'Firstname Lastname Surgery'. In the sentence of 'Known to Dr Firstname Lastname's Room…', the word of 'Room' was included as part of PERSON entity. These are errors from the perspective of strict NER but can be accepted in the de-identification task because they do not pose an identification risk.

2) Spelling mistakes (n=3). For example, 'Pro Lastname' and 'DrLastname' were not tagged correctly because of the spelling. The model could recognize the PERSON entities correctly when they were changed to 'Prof Lastname' and 'Dr Lastname'.

3) Non-tagged entities (n=13). Among these, seven entities were related to clinician names and six to patient names. The patterns of these entities were rare in our training set and performance could be improved by adding more training samples with the same patterns. Two entities did not have any context words around them.

All the false negative IDN entities were due to boundary mismatching. One false negative ADDRESS and two false negative PHONE entities were recognized as IDN entities. The rest of false negatives were not tagged by the ensemble model and almost all of them were related to clinician's phone numbers or addresses except one patient address.

The comparative results for performance of our model and two SOTA models on the 2014 i2b2 dataset are provided in Table 6. Our ensemble model achieved the highest F1 scores at 96.24% with strict entity matching and 98.64% with binary PII token-based matching, respectively.

Table 6. Comparation with two SOTA methods with our ensemble model on the i2b2 dataset. The best results for each metric appear in bold.

| System | Strict Entity | | | Binary PII Token | | |
|---|---|---|---|---|---|---|
| | Precision | Recall | F1 | Precision | Recall | F1 |
| Kim et al. [14] | **97.04** | 94.45 | 95.73 | 99.16 | 98.06 | 98.61 |
| Murugadoss et al. [18] | n/a | n/a | n/a | 97.90 | **99.20** | 98.50 |
| Our Ensemble | 96.90 | **95.59** | **96.24** | **99.43** | 97.86 | **98.64** |

We conducted an error analysis of our best ensemble model on the 2014 i2b2 test set as shown in Table S6. False positive and false negative errors were similar in type to those identified in the above error analysis. The errors found were either due to misclassifications from related entity categories (e.g., PATIENT versus DOCTOR and COUNTRY versus STATE) or boundary mismatches.

## 5. Discussion

In the present study, we developed and implemented an end-to-end framework for de-identification of Australian discharge summaries with three components: annotation, modelling, and de-identification. The framework achieved excellent performance on the test set of our corpus and obtained highly competitive results on the 2014 i2b2 de-identification dataset.

Our results show that an ensemble of different base-models which are trained using different deep learning algorithms performed better than the individual base-models. On our corpus, the F1 score of the best ensemble model was 99.16%, compared to the F1 score of 99.05% obtained from the best individual base-model. The difference was not statistically significant, however. The main reason might be that the discharge summaries in our corpus were highly homogeneous and had similar content structure and template on which the individual base-model already achieved good results. Meanwhile, our best ensemble model increased the strict entity F1 score by 0.87% from 95.73% to 96.24% on the 2014 i2b2 dataset, compared with Kim's model [14]. Further, our findings indicate that the proportion of text lines without PII entities in the training set affects the precision and recall scores of the trained models. In this study, we evaluated the models on two different training sets, namely, balanced, and imbalanced. The more text lines without PII entities in the training set (i.e., the greater the imbalance), the higher the precision score the models will achieve at the expense of recall. On the other hand, in a balanced dataset scenario where the number of text lines with and without PII entities are the same, the models achieve higher recall at the expense of precision. An ensemble approach that combines both models trained on balanced and imbalanced datasets produced the best overall precision-recall results in terms of F1 score. Hassanzadeh et al. [53], reported similar findings, where an ensemble classifier's F1 score was improved by up to 15% using balancing approaches on imbalanced datasets. Moreover, there were some semi-structured text lines which contained PII entities in the discharge summaries of our corpus such as "Patient: Firstname Lastname MRN: 123456 FIN 789012" and "Sex: Male DOB: 11-11-2025". The PII entities were "Firstname Lastname" as PERSON, "123456" and "789012" as IDN and "11-11-2025" as DOB. Generally, rule-based methods could be used to recognize these semi-structured entities but require manual effort to understand the whole dataset to create the rules. Our ensemble model perfectly extracted these patterns since there were no false positives or false negatives related to these semi-structured PII entities. All the experiments in this study were conducted on our Australian discharge summary corpus and the 2014 i2b2 (non-Australian) dataset which both contained hundreds of clinical documents. Ongoing evaluations, as part of future work, to assess the practical utility of our ensemble method, especially on larger datasets, will be pursued.

To investigate whether our ensemble method performed consistently well using different training and test sets, we conducted a 10-fold cross validation on our corpus. We first randomly shuffled the 500 discharge summaries in the pre-sampled training and development sets and split them equally into 10 folds. The training process was repeated 10 times. For each iteration, nine folds were used for training and validation and the remaining test fold was used to report performances. Our ensemble method achieved an average F1 score of 97.48% (±0.67%). Analysing the test fold errors revealed human annotation errors in the 500 annotated discharge summaries such as mislabelled PII categories and entity boundaries. The adjusted average F1 score, after updating the mislabelled annotations, was 98.98% (±0.48%), which was statistically similar to the F1 score (99.16%) from the final best ensemble model on the double annotated unseen test set as reported in Table 5. It shows the robustness of our ensemble method in selecting the best de-identification model. Table S7 provides the results from the individual models.

The best ensemble models on our corpus and the 2014 i2b2 dataset were combined using different base-models and ensemble methods. This demonstrated that there was not one ensemble method which could always perform best on different datasets. In our framework, we propose a set of ensemble models which were combined with majority voting and three stacking methods upon three groups of base-models: all the base-models, the three base-models with the best F1 scores and the three base-models with the best recall scores. When applying our framework on another specific dataset, we propose that the best ensemble model will be picked out of a set of ensemble models according to its F1 score.

This study demonstrated that the ensemble model can achieve good results even if only trained on hundreds of documents. We trained our models on only 500 discharge summaries and achieved a F1 score of 99.16% on 100 test discharge summaries. Similarly, we trained the models on 790 clinical notes and obtained a F1 score of 96.24% on the 2014 i2b2 dataset. This will facilitate application of our framework on any kind of clinical narrative documents because human annotation is costly and time-consuming.

The three components of our proposed end-to-end de-identification framework can form a learning loop, for ongoing improvement in the performance of the de-identification system, as shown in Fig. 5. We start from randomly selecting hundreds of raw documents as the input of the learning loop. Human annotators participate in the annotation component to tag the PII entities in the documents. The ensemble models are trained, and the best model is selected using the tagged documents in the modelling component. The de-identification component takes the best ensemble model and raw documents as input and outputs the PII-tagged documents and the de-identified documents. The PII-tagged documents can be fed back to the annotation component for annotators to correct incorrectly machine-assigned tags and generate more training data, presenting the potential to retrain the model to correctly identify entities that were initially non-tagged. The learning loop can accelerate the de-identification progress with less time and cost.

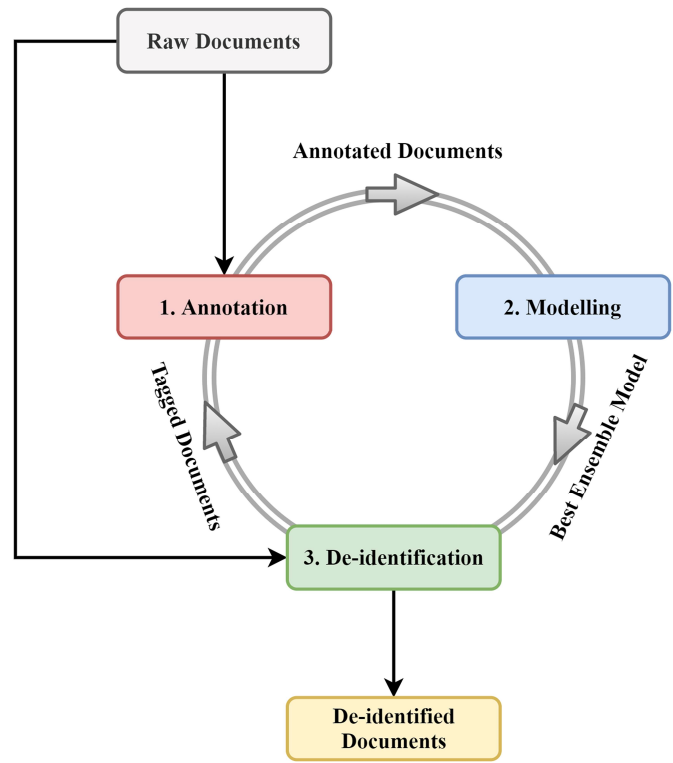

**Fig. 5.** The "learning loop" of the proposed end-to-end de-identification framework.

According to our error analysis, five of our false positives related to the inclusion of medical terminologies. That is because we used pre-trained word embeddings that did not capture the syntactic and semantic information of these terminologies. Reimers et al. [54] demonstrated that the word embeddings had a significant impact on the performance of deep learning NLP models. We will investigate some pre-trained medical domain-specific language representation models such as BioBERT [55] and BlueBERT [56] or training the word embedding on a research corpus that includes, for example, tagged lists of medical eponyms [57] and evaluate its performance as part of our future research. In addition, previous studies [19, 58, 59] have demonstrated that fine-tuning BERT models can achieve state-of-the-art performance on various downstream NLP tasks. This would be another promising future research area for de-identification tasks.

It is difficult to determine whether automated de-identified health data is safe to share and what level of performance can be relied on to permit automated de-identification systems. There is no industry-wide standard, but 95% is reasonable to be a rule-of-thumb for evaluating whether a de-identification system is reliable for de-identifying a dataset [12, 13]. However, the errors from de-identification systems must be carefully considered when applying them, especially for false negative errors which have the potential to threaten patient privacy.

## 6. Conclusion

In this study, we proposed an end-to-end framework for de-identifying Australian hospital discharge summaries. We trained six base-models and fine-tuned the hyperparameters on balanced and imbalanced datasets using BiLSTM+CRF and CNN algorithms. A set of ensemble models were aggregated upon three groups of base-models with majority voting and stacking methods. We compared these models using the metrics of strict micro F1 score on the test set. The ensemble model combined using stacking SVM method on the three base-models with the best F1 scores performed the best with a F1 score of 99.16%. We used the best ensemble model to tag all the PII entities and replace them with relevant surrogates in hospital discharge summaries. Meanwhile, we evaluated the robustness of the modelling process on the 2014 i2b2 dataset. Our ensemble model generated using the majority voting method on all the base-models achieved the highest F1 scores compared to two SOTA methods.

Our end-to-end de-identification framework can be applied to discharge summaries from other hospitals and health systems, and to other types of narrative documents like progress notes. Moreover, the learning loop of the framework will keep improving the accuracy of de-identification. We believe that our end-to-end framework can provide a robust solution for de-identifying and sharing clinical narrative text safely.

## Ethics approval and consent to participate

## CRediT authorship contribution statement

**Leibo Liu:** Conceptualization, Methodology, Software, Model Building, Evaluation, Writing - original draft. **Oscar Perez-Concha:** Conceptualization, Methodology, Supervision, Writing - review & editing. **Anthony Nguyen:** Methodology, Supervision, Writing - review & editing. **Vicki Bennett:** Methodology, Supervision, Writing - review & editing. **Louisa Jorm:** Conceptualization, Methodology, Supervision, Writing - review & editing.

## Declaration of Competing Interest

The authors declare that they have no known competing financial interests or personal relationships that could have appeared to influence the work reported in this paper.

**Funding**

This study was supported by the Australian government and the Commonwealth Industrial and Scientific Research Organisation (CSIRO) through Australian Government Research Training Program scholarship and CSIRO top up scholarship.